\documentclass[letterpaper, 10pt, conference]{ieeeconf} 

\IEEEoverridecommandlockouts                    
                                                          
\usepackage{multirow}

\usepackage{algpseudocode}

\usepackage{rotating}
\usepackage{color}
\usepackage{subfigure}
\usepackage{algorithm}
\usepackage[justification=centering]{caption}
\usepackage{braket}
\usepackage{amssymb}
\usepackage{cancel}

\newcommand\hcancel[2][black]{\setbox0=\hbox{$#2$}%
\rlap{\raisebox{.45\ht0}{\textcolor{#1}{\rule{\wd0}{1pt}}}}#2} 

\definecolor{darkgreen}{rgb}{0,0.8,0}

\title{\LARGE \bf
What is (missing or wrong) in the scene? A Hybrid Deep Boltzmann Machine For Contextualized Scene Modeling
}

\author{\.{I}lker Bozcan$^{1}$, Ya\~gmur Oymak$^{1}$, \.{I}dil Zeynep Alemdar$^{1}$ and Sinan Kalkan$^{1}$
\thanks{$^{1}$All authors are with the KOVAN Research Lab at the Department of Computer Engineering, Middle East Technical University, Ankara, Turkey
        {\tt\small \{ilker.bozcan, yagmur.oymak, idil.alemdar, skalkan\}@metu.edu.tr}}%
}

\begin{document}

\maketitle
\thispagestyle{empty}
\pagestyle{empty}

\begin{abstract}
Scene models allow robots to reason about what is in the scene, what else should be in it, and what should not be in it. In this paper, we propose a hybrid Boltzmann Machine (BM) for scene modeling where relations between objects are integrated. To be able to do that, we extend BM to include tri-way edges between visible (object) nodes and make the network to share the relations across different objects. We evaluate our method against several baseline models (Deep Boltzmann Machines, and Restricted Boltzmann Machines) on a scene classification dataset, and show that it performs better in several scene reasoning tasks.
\end{abstract}

\section{Introduction}

Modeling (representing) their environments is crucial for cognitive as well as artificial agents. For a robot, scene modeling pertains to representing a scene in such a way that the robot can reason about the scene and the objects in it in an efficient manner. A scene model should allow for the robot to check, for example, (i) whether there is a a certain object in the scene and where it is, or (ii) whether it is in the right place in the scene or (iii) whether there is something redundant in the scene to be moved to some place else.

Although there are many studies on scene modeling in robotics and computer vision (e.g., \cite{CelikkanatConceptWeb2014,CelikkanatContext2014,li2017context,anand2013contextually}), to the best of our knowledge, ours is the first that uses a multi-way Boltzmann Machines for the task. 

\begin{figure}
\centerline{
	\subfigure[]{
        \label{fig:triway_BM}
    	\includegraphics[width=0.49\textwidth]{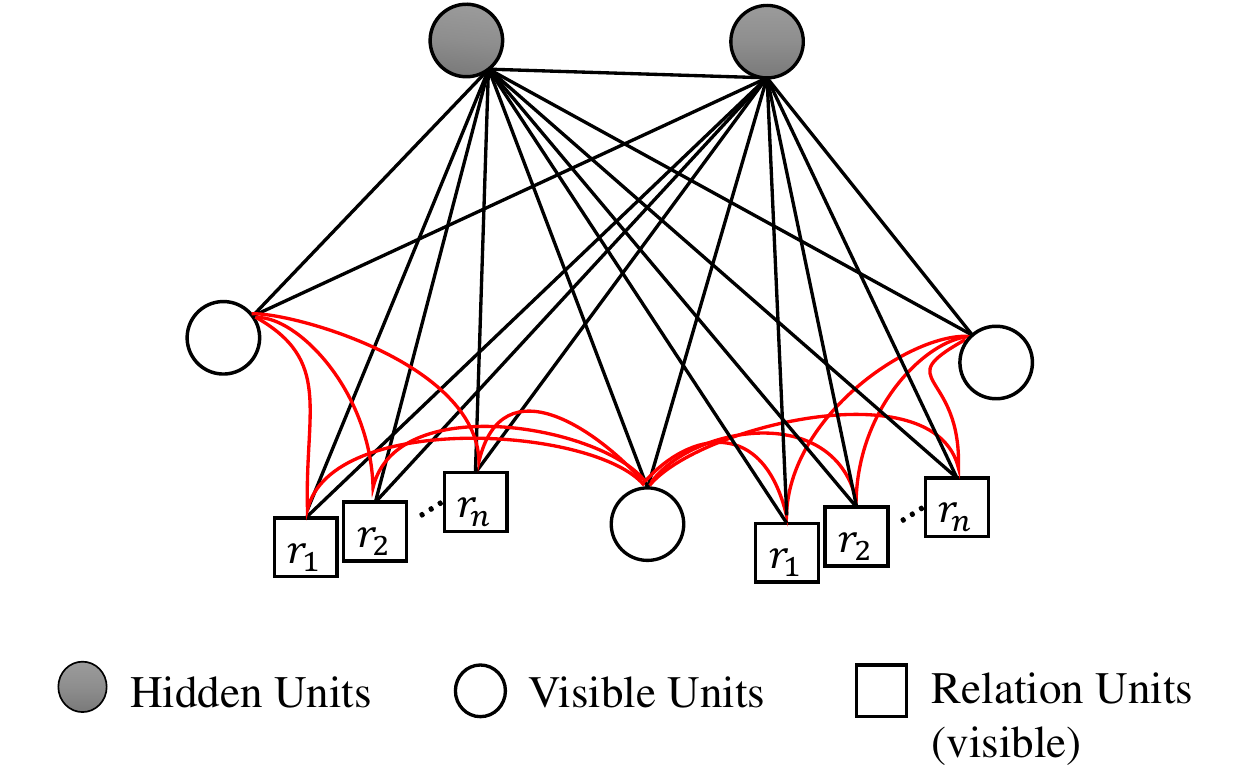}
    }
}
\centerline{
	\subfigure[]{
    	\includegraphics[width=0.49\textwidth]{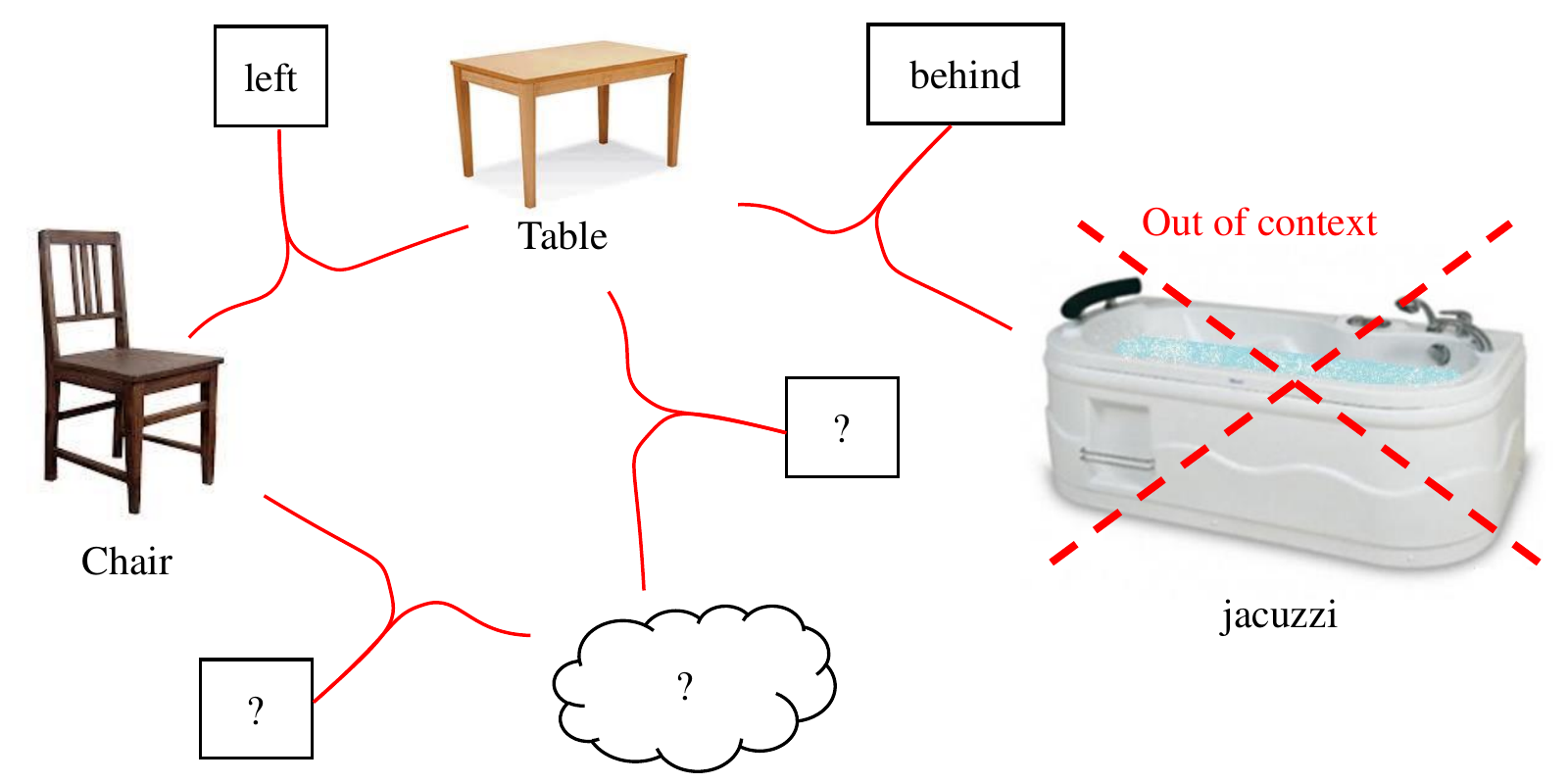}
    }
}
\caption{(a) An overview of the proposed hybrid tri-way Boltzmann Machine, where the tri-way edges are shown in red, and (b) some examples for what it can provide to a robot (given some incomplete or wrong observations from the environment). [Best viewed in color] \label{fig:overview}}
\end{figure}

Boltzmann Machines (BM) \cite{ackley1985learning} are stochastic generative models that offer many benefits for various modeling problems. The benefits of Boltzmann Machines include (among others) the presence of latent nodes, which function as context variables modulating the object activations, the ease to extend with the requirements of scene modeling, and its generative capability. Although BM existed beforehand, they became popular again with extensions to deep architectures or restricted connections (i.e., Restricted Boltzmann Machines). 

\subsection{Related Work}

\textbf{Scene Modeling:} Many models have been proposed for scene modeling in computer vision and robotics using probabilistic models such as Markov or Conditional Random Fields \cite{CelikkanatConceptWeb2014,CelikkanatContext2014,anand2013contextually,lin2013holistic}, Bayesian Networks \cite{li2017context,sheikh2005bayesian}, Latent Dirichlet Allocation variants \cite{wang2008spatial, Philbin08a}, Dirichlet and Beta processes \cite{joho2013nonparametric}, chain-graphs \cite{pronobis2012large}, predicate logic \cite{mastrogiovanni2011robots,hwang2006ontology}, and Scene Graphs \cite{blumenthal2014towards}. There have also been many attempts for ontology-based scene modeling where objects and various types of relations are modeled \cite{hwang2006ontology,saxena2014robobrain,tenorth2009knowrob}.

Among these, \cite{CelikkanatConceptWeb2014,CelikkanatContext2014,li2017context,anand2013contextually} use context either in representing the scene or solving a task using the scene model for a robotics problem. These studies model context via local interactions between visible variables, except for \cite{CelikkanatContext2014} who proposed using Latent Dirichlet Allocation for modeling context.

\textbf{Relation Estimation and Reasoning:}
Early studies on integrating relations into scene modeling and analysis tasks were rule-based. These approaches defined relations using rules based on 2D/3D distances between objects, e.g., \cite{stopp1994utilizing}. With advances in probabilistic graphical modeling, many approaches used models such as Markov Random Fields \cite{anand2013contextually,celikkanat2015integrating}, Conditional Random Fields \cite{lin2013holistic}, Implicit Shape Models \cite{meissner2013recognizing}, latent generative models \cite{joho2013nonparametric}. Many studies also proposed formulating relation detection as a classification problem, e.g., using logistic regression \cite{guadarrama2013grounding}, and deep learning \cite{johnson2016clevr}.

\subsection{Contributions}

The main contributions of our work are the following:
\begin{itemize}
\item \textbf{Deep Boltzmann Machines for Scene Modeling:} We use Deep Boltzmann Machines (DBM) for modeling a scene in terms of objects and the relations between the objects. To the best of our knowledge, this is the first study that uses DBM with relations for the task.
\item \textbf{A Hybrid Triway Model - DBM with relations:} Adding relations to DBM is not straightforward since there may be different relations between objects and the same relations between different objects should represent the same thing. This leads to two extensions: (i) Tri-way nodes to represent relations in the DBM, (ii) Weight-sharing between the weights of relation nodes to enforce relations between different objects to represent the same relations.
\end{itemize}

We evaluate our extended DBM model on many practical robot problems: Determining (i) what is missing in a scene, (ii) relations between objects, (iii) what should not be in a scene,  (iv) random scene generation given some objects or relations from the to-be-generated scene. We compare our model (Triway BM) against DBM \cite{salakhutdinov2009deep} with 2-way relations (GBM), and Restricted Boltzmann Machines (RBM) \cite{salakhutdinov2007restricted}.

\section{Background: General, Restricted and Higher-order Boltzmann Machines}


\begin{figure} 
\centerline{
\includegraphics[width=0.49\textwidth]{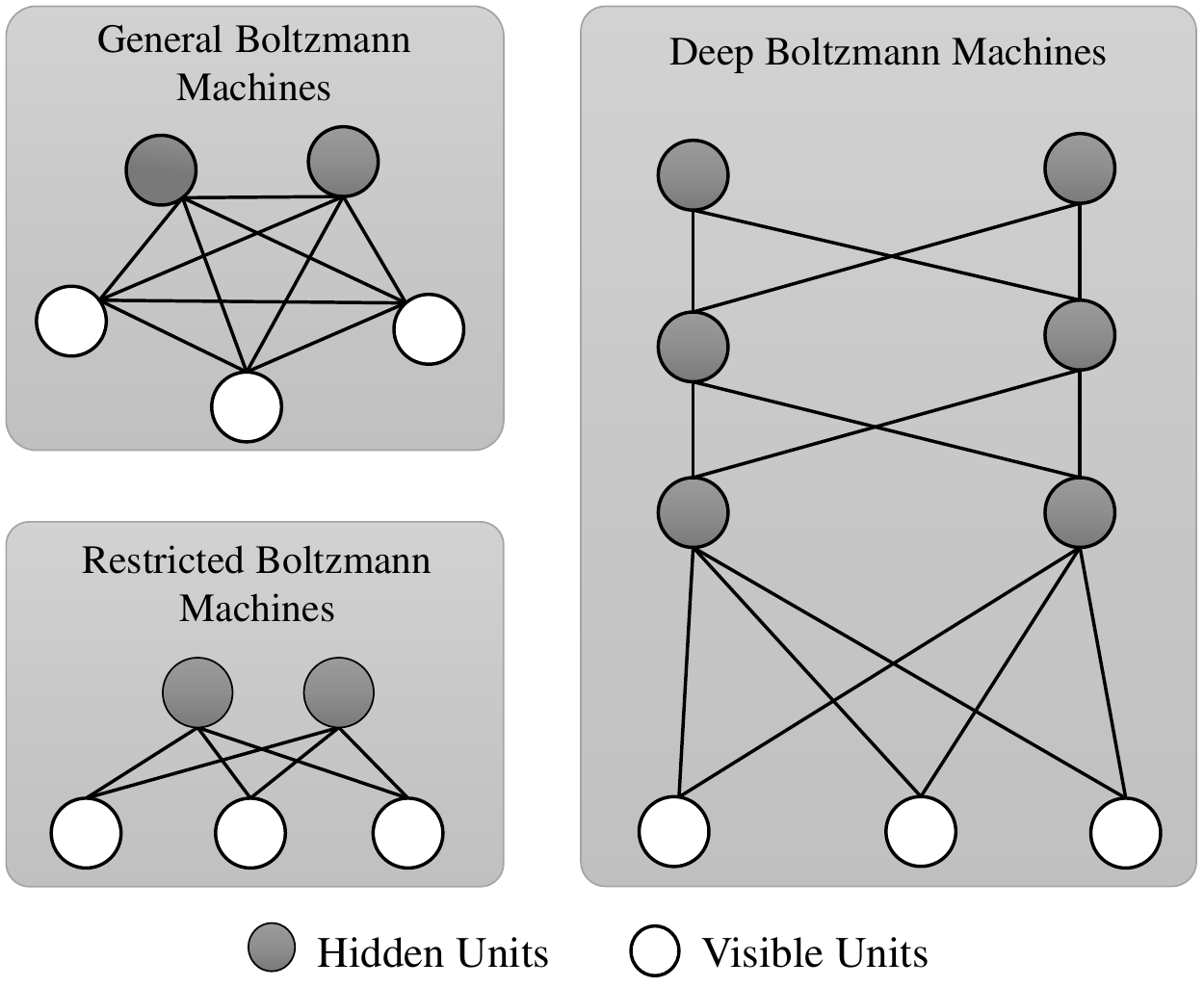}
}
\caption{A schematic comparison of Boltzmann Machines, Restricted Boltzmann Machines and Deep Boltzmann Machines. \label{fig:all_bm}}
\end{figure}

A Boltzmann Machine (BM)\footnote{Although this is textbook material, it is essential for us to be able to describe our extensions.} \cite{ackley1985learning} is a graphical model composed of visible nodes $\mathbf{v}=\{v_i\}_{i=1}^{V} \subset \{0,1\}^V$ and hidden nodes $\mathbf{h}=\{h_i\}_{i=1}^{H} \subset \{0,1\}^H$ -- see also Figure \ref{fig:all_bm}. In a BM, hidden nodes are connected to other hidden nodes with bi-directional weights, $W^{hh}= \{w^{hh}_{ij}\}^{H\times H}$; visible nodes to other visible nodes with $W^{vv}=\{w^{vv}_{ij}\}^{V\times V}$; and hidden nodes to visible nodes with $W^{hv}=\{w^{hv}_{ij}\}^{H\times V}$. With these connections, a BM tries to obtain an estimation of $p(\mathbf{v})=\sum_{\mathbf{h}} p(\mathbf{v}|\mathbf{h})p(\mathbf(h)) $ from a sample of the training data. 

For a BM, one can define a scalar, representing the negative harmony between the nodes given current weights:
\begin{equation}
E(\textbf{v}, \textbf{h})=-\sum_{i<j}v_i w^{vv}_{ij} v_j
              -\sum_{i<j}h_i w^{hh}_{ij} h_j
              -\sum_{i<j}h_i w^{hv}_{ij} v_j. \label{eqn:energy_BM}
\end{equation}
BM is inspired from physical systems which favor states with lower energies, and therefore, the probability of being in a certain state (i.e., $\{\mathbf{v}, \mathbf{h}\}$) is linked to the energy of the system via Boltzmann distribution:
\begin{equation}
p(\mathbf{v}, \mathbf{h})= \frac{1}{Z} \exp(-E(\textbf{v}, \textbf{h})),
\end{equation}
{\noindent}where $Z=\sum_{\textbf{v'}, \textbf{h'}} E(\textbf{v'}, \textbf{h'})$ is called the partitioning function. Since the partitioning function is intractable to calculate for real problems, $p(\mathbf{v}, \mathbf{h})$ is iteratively estimated by stochastically activating nodes in the network with probability based on the change in the energy of the system after the update:
\begin{equation}
p(x=1) = \frac{1}{1+e^{\Delta E_x/T}},
\end{equation}
{\noindent}where $x$ is a visible or a hidden node; $\Delta E_x$ is the change in energy of the system if $x$ is turned on; and $T$ is the temperature of the system, gradually decreased (annealed) to a low value, controlling how stochastic the updates are.

Since training is rather slow and limiting in BM, its restricted version (Restricted Boltzmann Machines) with only connections between hidden and visible nodes have been proposed \cite{salakhutdinov2007restricted}. In a Deep Boltzmann Machine \cite{salakhutdinov2009deep}, on the other hand, there are layers of hidden nodes. See Figure \ref{fig:all_bm} for a schematic comparison of the alternative models.

Some problems require the edges to combine more than two nodes at once, which have led to the Higher-order Boltzmann Machines (HBM) \cite{sejnowski1986higher}. With HBM, one can introduce edges of any order to link multiple nodes together.

\subsection{Training a Boltzmann Machine}

Training a BM minimizes the Kullback-Leibler divergence between $p^+(\textbf{v})$, the distribution over $\mathbf{v}$ when data is clamped on the visible nodes (called the positive phase), and $p^-(\mathbf{v})$, the distribution obtained when the network is run freely (called the negative phase). Taking the gradient of the divergence with respect to the weights leads to the following update rule:
\begin{equation}
w_{ij} \leftarrow w_{ij} - \alpha (p^+_{ij} - p^-_{ij}),
\end{equation}
{\noindent}where $p^+_{ij}$ and $p^-_{ij}$ are the expected joint activations of nodes $s_i$ and $s_j$ during the positive phase and the negative phase, respectively; and $\alpha$ is a learning rate.

\section{A Triway Hybrid Boltzmann Machine for Scene Modeling}

As shown in Figure \ref{fig:triway_BM}, we extend Boltzmann Machines by adding relational (visible) nodes $r_i \in \mathbb{B}$ that (i) are shared across objects, and (ii) link two objects together with a single tri-way edge. In other words, a relation $r_i$ connects two objects, $v_j$ and $v_k$ with a weight $w^r_{ijk}$. The overall energy of the hybrid BM then is updated as follows:
\begin{eqnarray}
E(\textbf{v}, \textbf{h}, \textcolor{red}{\textbf{r}}) & = & 
  			  \hcancel[red]{-\sum_{i<j}v_i w^{vv}_{ij} v_j}
              -\sum_{i<j}h_i w^{hh}_{ij} h_j \nonumber \\
         & &  -\sum_{i<j}h_i w^{hv}_{ij} v_j  \nonumber \\
         & &  \textcolor{red}{-\sum_{i,j,k} w^{r}_{ijk} r_i v_j v_k}
              \textcolor{red}{-\sum_{i<j} r_i w^{rh}_{ij} h_j} , \label{eqn:energy_hybridBM}
\end{eqnarray}
{\noindent}where the changes compared to the energy definition in Equation \ref{eqn:energy_BM} are highlighted in red. Note that the definition in Equation \ref{eqn:energy_hybridBM} uses tri-way edges with the relation nodes, and that relations (in fact, weights of relations) are shared across objects. 

Weight-sharing suggests that, e.g., a \textit{left} relation between $v_i$ and $v_j$, and a \textit{left} relation between $v_k$ and $v_l$ ($i\neq j \neq k \neq l$) represent the same relation. In order to do that, the gradients on the weights of relation $r_i$ that is coming from all pairs of objects in the scene are aggregated:
\begin{equation}
\Delta w_{ijk} = \sum_{s_j,s_k\ \in\ U_{r_i}} (p^+_{ijk} - p^-_{ijk}),
\end{equation}
{\noindent}where $U_{r_i}$ is the set of object tuples connected by relation $r_i$.


\subsection{Training and Inference}

In order to make training faster, we dropped the connections between the hidden neurons.

For training our Triway Hybrid BM, in the positive phase, as usual, we clamp the visible units with the objects, and the relations between the objects and calculate $p^+$ for any edge in the network. 

In negative phase, firstly, object units are sampled with a two-step Gibbs sampling by using activation of hidden units and relation units. In this way, relation units also contribute to activation of object units, in addition to the hidden units. Then, the relation units are sampled from recently sampled object units and hidden units. We calculate $p^-$ for any edge in the network in these two steps.

For training the networks, we used gradient descent with a batch size  of 32 with early stopping (i.e., training process is finished when validation accuracy begins to decrease). Learning rate and temperature are empirically set to 0.5 and 1 respectively and 2 hidden layers is used that the bottom layer has 200 hidden units and the top layer has 100 hidden layers. 

For inference, we use Gibbs sampling \cite{geman1984stochastic} that is a Monte Carlo Markov Chain (MCMC) method to approximate true data distribution. We prefer a MCMC method over variational inference since our dataset is relatively small (totaling 3,485 samples) and input vectors are too sparse (i.e. slight number of relation nodes are active). Therefore, we need precise inferences that MCMC methods can guarantee but variational inference cannot \cite{blei2017variational}.

\subsection{Dataset Collection}

There are two datasets with labeled spatial relations \cite{golland2010game,johnson2016clevr}. However, both datasets are simulated, and therefore, we collected a real dataset with relations.


We use a 3,485 (the ones acquired with newest depth sensors) of the SUNRGBD dataset \cite{song2015sun}. Misspelled and redundant object labels were merged. In this way, total numbers of objects are reduced to 417. We extended the original dataset by adding eight spatial (left, right, front, behind, on-top, under, above, below) relations among annotated objects manually. All object pairs are considered as a relation. Therefore, a total number of relations that can be estimated is $8 \times 417\times 417=1,391,112$. 

Let us use $D=\{S_1,...,S_{3485}\}$, where $S_i$ denotes $i^{th}$ sample, to denote the dataset. $S_i$ has a vector form that represents the presence of objects and relations among them in the scene. Active objects and relations have value 1, otherwise 0. Opposite relations (ex: left and right) can be represented in one relation in BMs since if object $a$ is left of object $b$ then object $b$ is right of object $a$. As a result, each sample is represented by a binary vector that has length 695,973 $(417+4\times 417\times 417)$. 

There are 33 indoor scene types (kitchen, dining room etc.) that robots can be used for variety of tasks instead of humans in dataset. We split dataset into three: $60\%$ for training, $30\%$ for testing and $10\%$ for validation during training. All sets include samples from each scene category.

\section{Experiments and Results}

In this section, we evaluate and compare the methods on several tasks.

\subsection{Network Training Performance}

We calculated an error on difference between original data that are clamped to visible units and reconstructed visible states that are sampled in negative phase and observed how it changes during training:
\begin{equation}
E_{train} = \frac{1}{|\mathbf{V}|}\sum_{\mathbf{v}\in\mathbf{V}}\sum_{j}(p(v_j^{+})-p(v_j^{-}))^2 ,
\label{train_error_equation}
\end{equation}
{\noindent}where $p(v^{+})$ is probability of activation of original data that is clamped to visible units at the beginning of positive phase. $P(v^{-})$ is probability of activation of visible nodes at the end of the negative phase. The cumulative sum over all samples is normalized with the total number of samples ($|\mathbf{V}|$).

\begin{figure}
\centerline{
    	\includegraphics[width=0.49\textwidth]{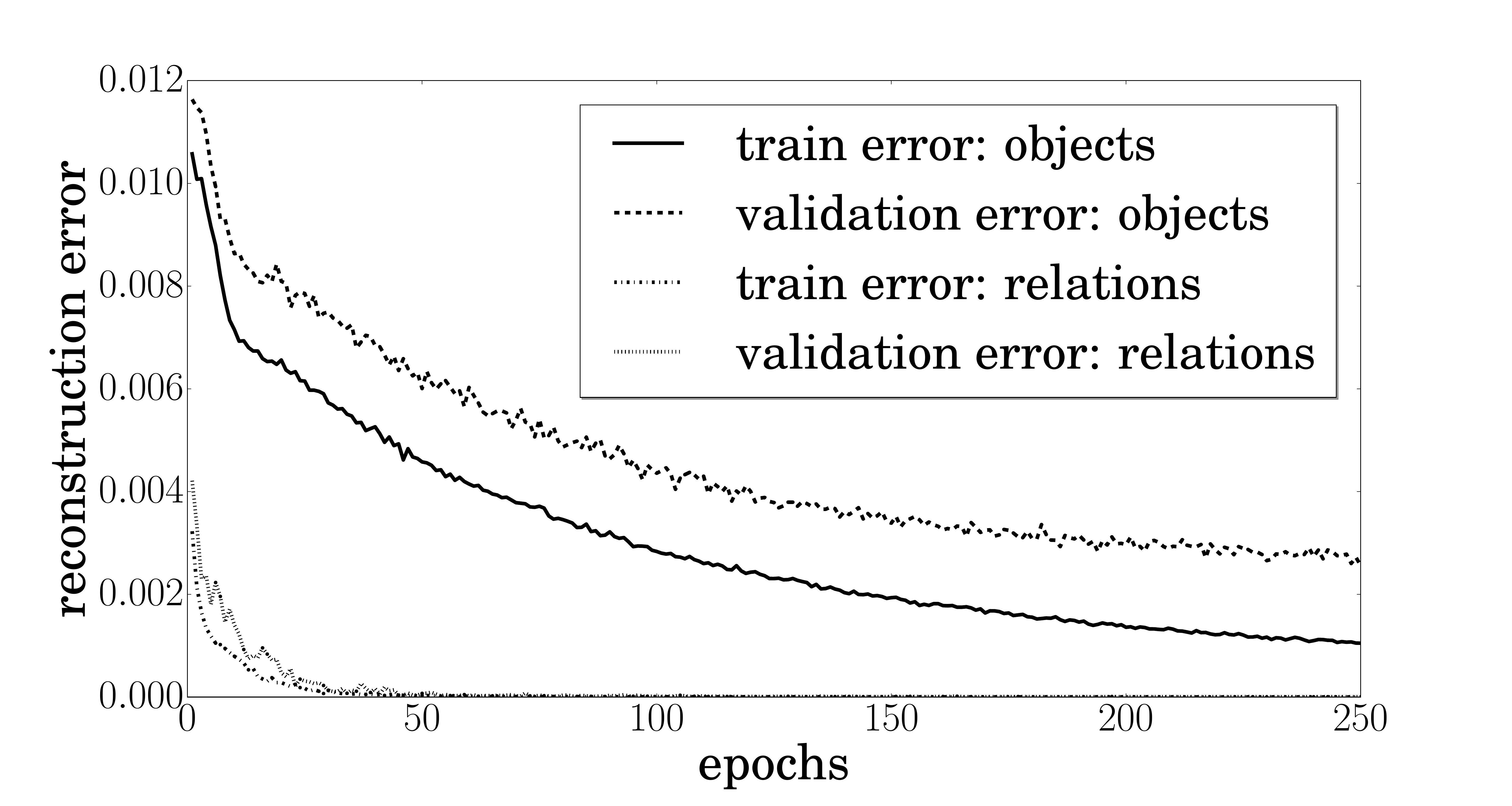}
}
\caption{Reconstruction error vs. epochs plot during Triway BM training. \label{fig:loss}}
\end{figure}

We look at the error separately for the objects and the relations, as shown in Figure \ref{fig:loss}. We see that the network consistently decreases the error, and learns to represent objects and the relations between them. However, it somehow learns relations much faster. This difference is because the space of all possible relations is much larger than the objects set, and very sparse; therefore, the network quickly learns to estimate 0 (zero) for relations, which leads to a sudden decrease in the loss.

\subsection{Task 1: Relation Estimation}

Our model can estimate possible relations among active concepts in the scene. For testing, active objects on the scene are clamped to visible object nodes and the model is let loose. Initially, model sees the environment in terms of ``bag of objects'' and samples hidden units (i.e., context). The context is determined by using objects on the scene without relations among them. Then, the relation nodes are sampled from contexts and objects. For this task, we define accuracy as the percentage of relations correctly estimated with respect to the labeled relations in the test dataset.
 
\begin{table}[hbt]
\caption{Performance (accuracy) of the methods on estimating relations between a given set of objects  (Task 1). Higher is better. \label{tbl:relationestimation}}
\centering
\begin{tabular}{|cccc|}\hline                                                                 
RBM  & GBM  & Triway BM & Chance level \\ \hline \hline
5.60\% &    14.18\%  & \textbf{23.35\%} & $1.43\times10^{-4}$\%  \\ \hline
\end{tabular}
\end{table}

We evaluate this task with RBM, General BM and our Hybrid model, as shown in Table \ref{tbl:relationestimation}. We see that our model provides highest accuracy. We do not consider inactive relation nodes in original data since the network has already learned which relation nodes should be inactive.

\begin{figure}
\centerline{  
	\subfigure[]{
    	\label{fig:task1a}
    	\includegraphics[width=0.45\textwidth]{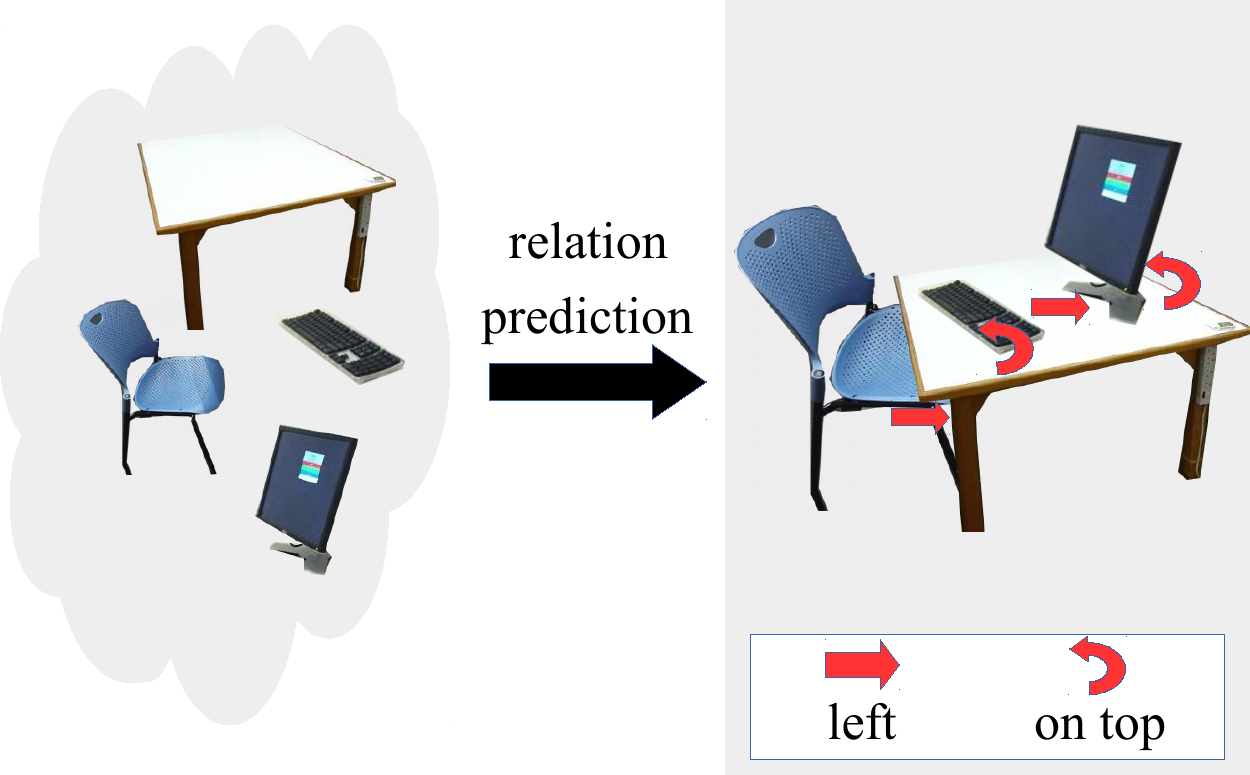}
        }
}
\vspace*{0.5cm}
\centerline{  
	\subfigure[]{
    	\label{fig:task1b}
    	\includegraphics[width=0.45\textwidth]{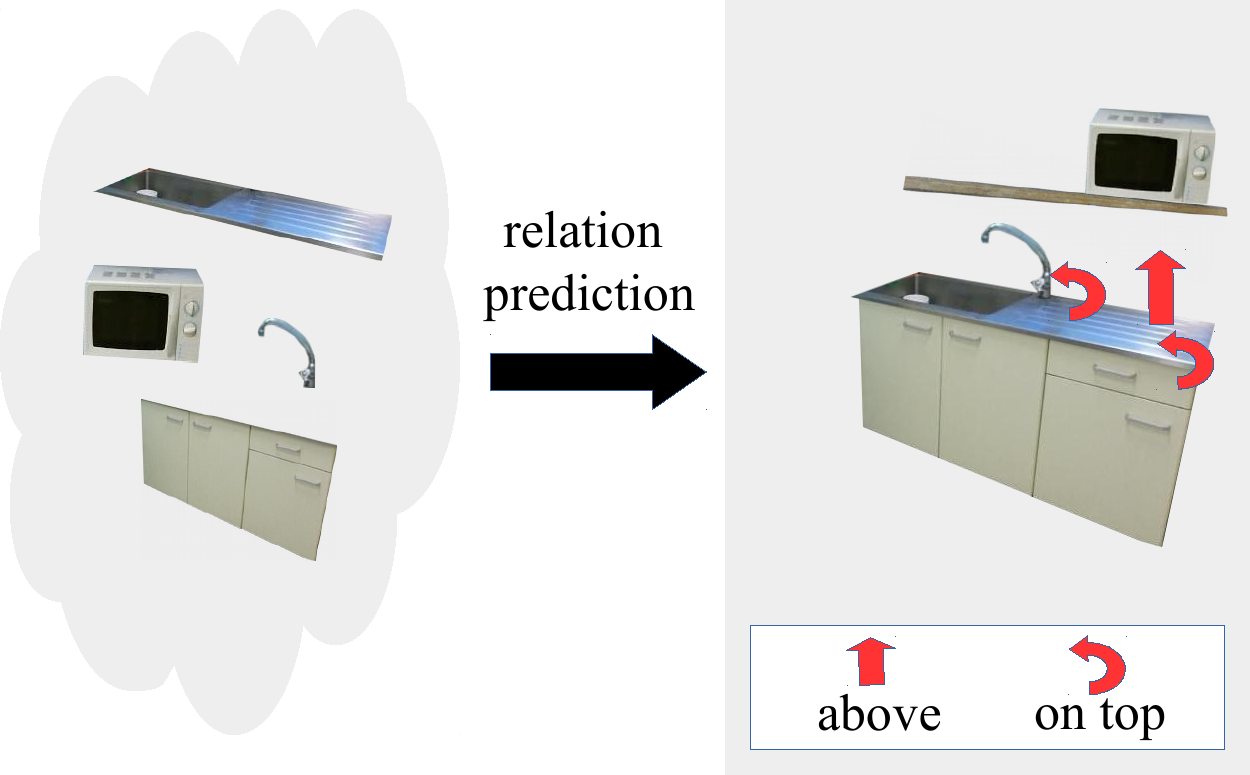}
        }
}
\caption{Some example relations estimated by our hybrid model for given sets of objects  (Task 1). Only a subset of the relations is shown for the sake of visibility. \label{fig:relationestimation}}
\end{figure}

We provide some visual examples in Figure \ref{fig:relationestimation}, where we see that our model nicely finds out how to place a set of objects together. The chance level of activation of one relation node is $\frac{1}{\mbox{\# of relation nodes}} \approx 1.43\times10^{-6}$. 

\subsection{Task 2: What is missing in the scene?}

In this task, we randomly de-activate an object from the scene and expect the network to find the missing object. For this task, we define accuracy as the percentage of the missing objects found correctly in the reconstructed sample. 

\begin{figure}
\centerline{
	\subfigure[]{
    	\label{fig:task2a}
    	\includegraphics[width=0.49\textwidth]{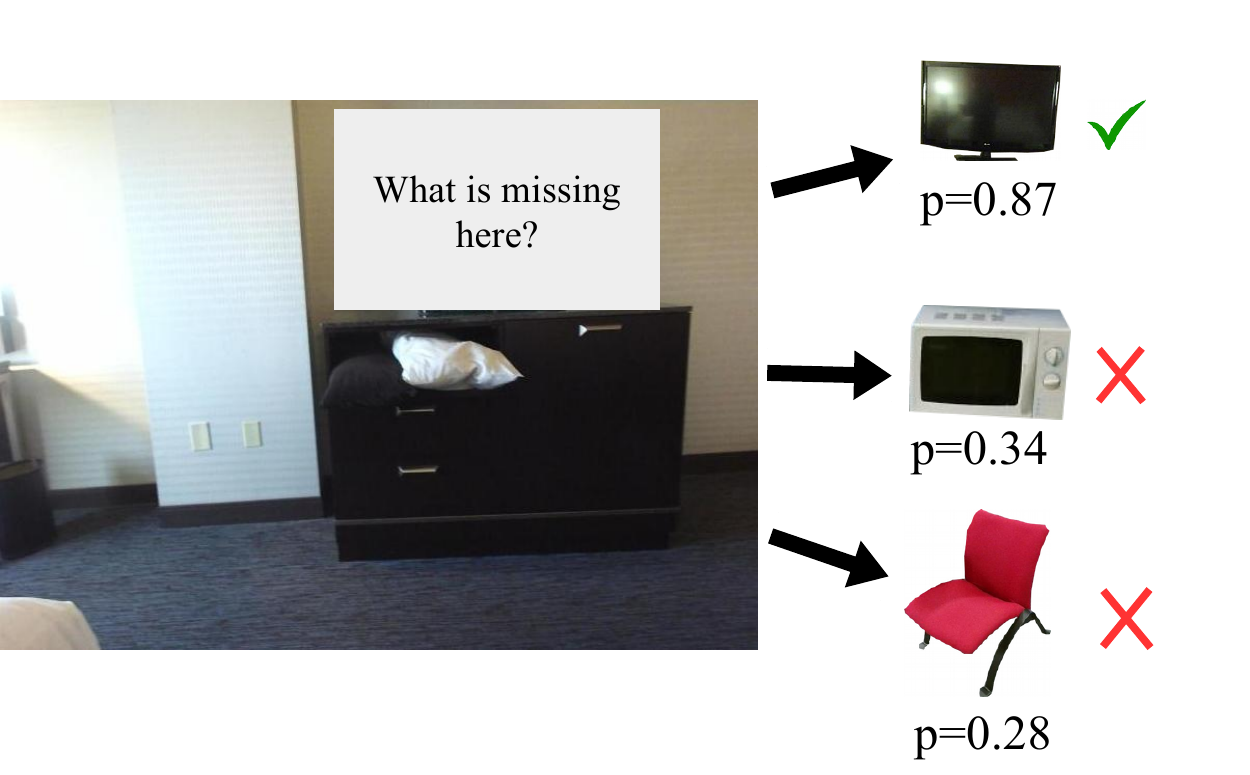}
        }
}
\centerline{ 
	\subfigure[]{
    	\label{fig:task2b}
    	\includegraphics[width=0.49\textwidth]{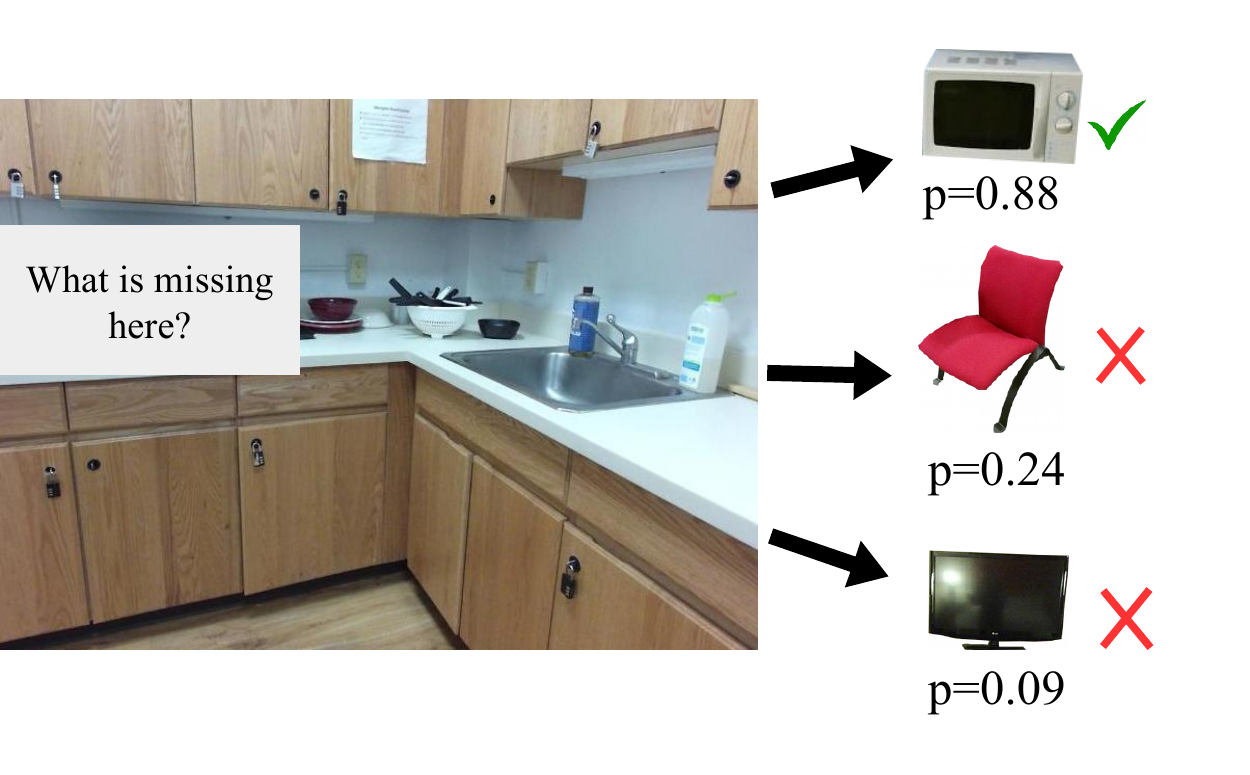}
    }
}
\caption{Some examples illustrating the performance of our hybrid model on finding a missing object in a scene  (Task 2). \label{fig:whatismissing}}
\end{figure}

As shown in Table \ref{tbl:whatismissing}, our hybrid model performs better than RBM and GBM. See also Figure \ref{fig:whatismissing}, which shows some visual examples for most likely objects found for a target position in the scene.

\begin{table}[H]
\centering
\caption{Performance (accuracy) of the methods on finding missing object in the scene  (Task 2). Higher is better.}
\label{tbl:whatismissing}
  \begin{tabular}{|cccc|}\hline 
  RBM  & GBM  & Triway BM & Chance level \\ \hline \hline               
  35.12\%   &  {40.94\%}   & \textbf{43.28}\%     &   $5.75\times10^{-6}$\%           \\ \hline
  \end{tabular}
\end{table}

\subsection{Task 3: What is out of context in the scene?}

\begin{figure}
\centerline{
	\subfigure[]{
    	\label{fig:task3a}
    	\includegraphics[width=0.49\textwidth]{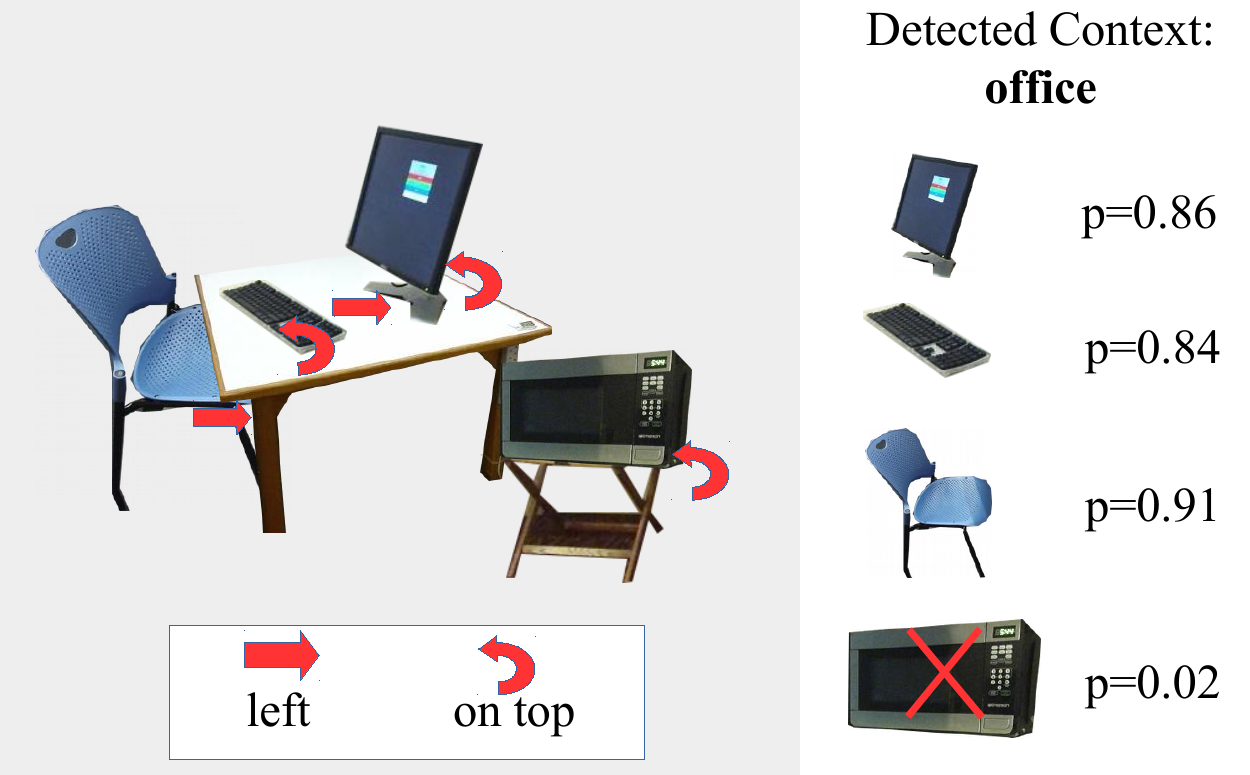}
        }
    }
\centerline{
	\subfigure[]{
    	\label{fig:task3b}
    	\includegraphics[width=0.49\textwidth]{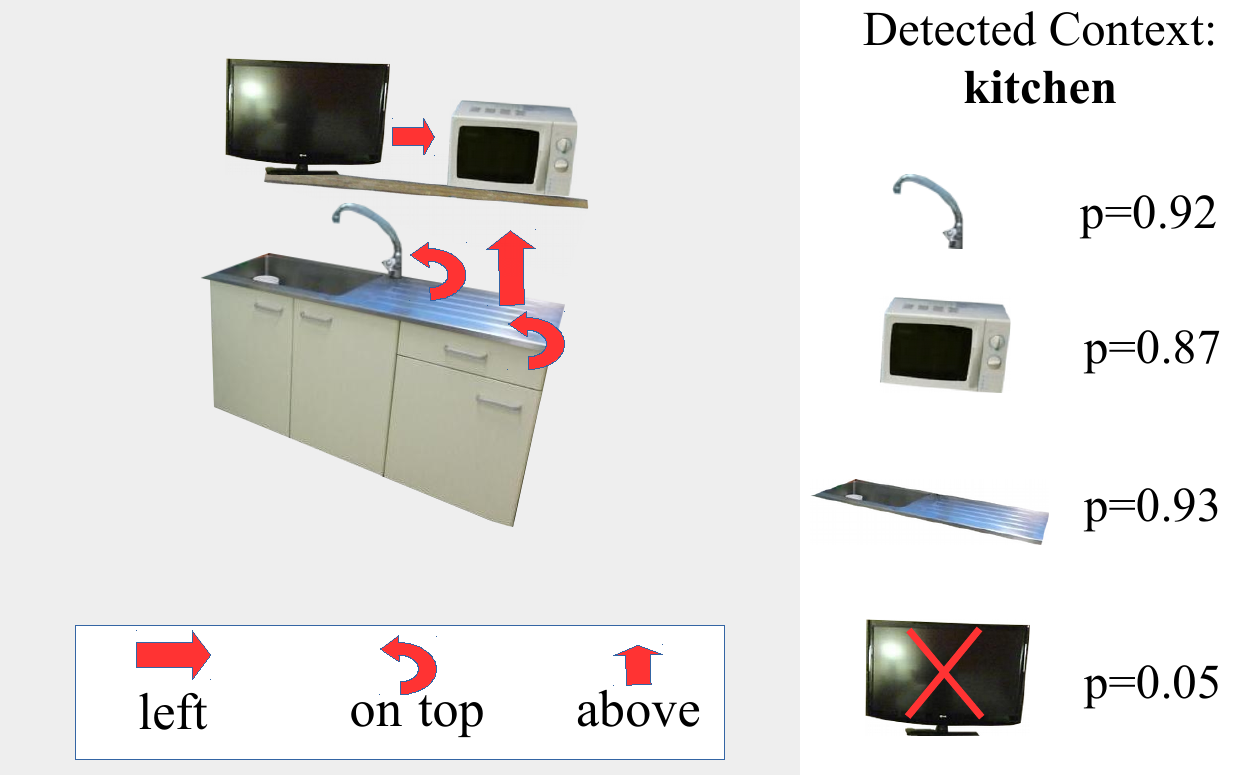}
    }
}
\caption{Some examples illustrating the performance of our hybrid model on finding what is out of context in a scene  (Task 3). \label{fig:whatisextra}}
\end{figure}

Next, we evaluate how good the methods find an object that is out of context in the scene. For this end, we select scenes,  remove randomly an object and  randomly add  another object not in the scene. Of course, during this process, the network might disrupt other objects in $\mathbf{v}$ as well. To take this into account, for this task, we define the error measure for a sample as the number of objects that are incorrectly sampled or changed. Let $\mathbf{v}^{+x}$ be the current scene representation, where $x$ is the randomly selected active object, and $\mathbf{v}^{-x}$ be the scene representation, where $x$ is removed (set to zero). After $\mathbf{v}^{-x}$ is clamped, the network settles to $\mathbf{v}^{?x}$, where object $x$ is hopefully recovered, but there might be other unwanted changes on nodes other than $x$. We can define our measure formally as follows:
\begin{equation}
 \textrm{Task 3 error measure} = \frac{1}{|\mathbf{v}|\times|\mathbf{V}|}\sum_{\mathbf{v}^{+x} \in \mathbf{V}} \sum_i \textrm{abs}(\mathbf{v}^{+x}_i - \mathbf{v}^{?x}),
\end{equation}
{\noindent}where $\textrm(abs(\cdot))$ is the absolute value function. Table \ref{tbl:whatisextra} compares the methods, and shows that our hybrid model produces lowest error. See also Figure \ref{fig:whatisextra} for some visual examples.

In this task, models may tend to give higher contextual importance to particular objects for different scenes (i.e. ``dishwasher" is a dominant object for the ``kitchen" context and provides higher contextual information than a ``chair" object for the ``kitchen" context). Therefore, they can remove objects that have lower contextual information and corrupt original input data.

\begin{table}
\centering
\caption{Performance (error) of the methods on finding what is out of context in the scene (Task 3). Lower is better. \label{tbl:whatisextra}}
  \begin{tabular}{|cccc|} \hline     
     RBM  & GBM  & Triway BM & Chance level \\ \hline \hline               
     0.6446  &  0.1404    &  \textbf{0.0789}  & 0.5              \\ \hline
  \end{tabular}
\end{table}

\subsection{Task 4: Generate a scene}

\begin{figure}

\centerline{
	\subfigure[]{
        	\label{fig:task4a}
        	\includegraphics[width=0.49\textwidth]{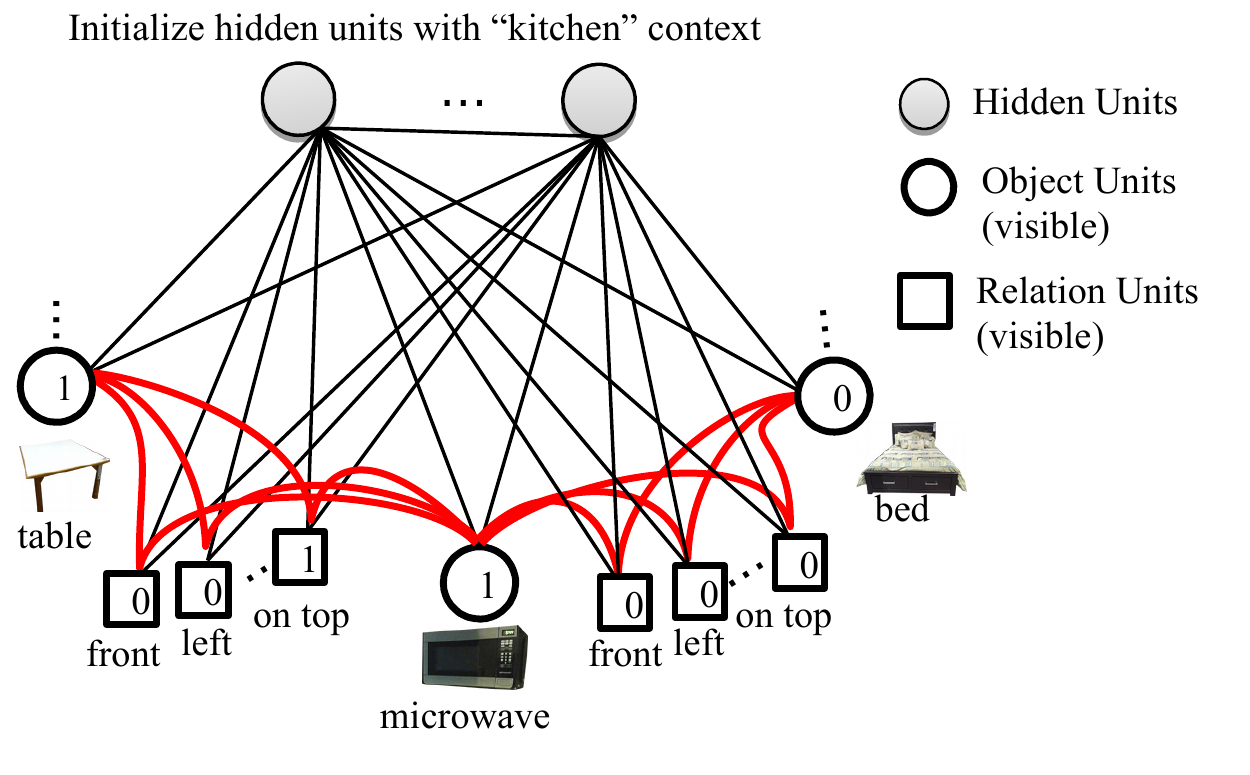}
    }
}

\centerline{
  	\subfigure[]{
    	\label{fig:task4b}
    	\includegraphics[width=0.49\textwidth] {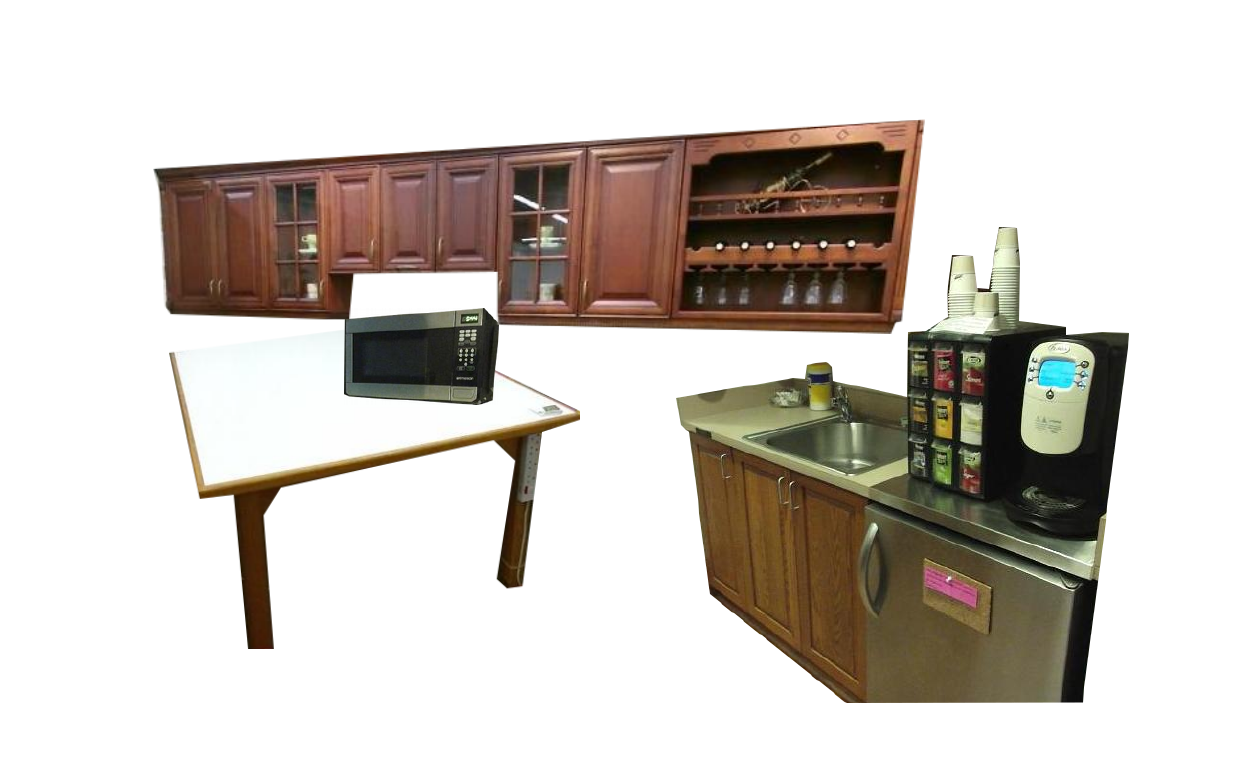}
    }
}

\caption{An example illustrating scene generation capability of our hybrid model. (a) When a context (hidden node) is activated, (b) active nodes in the sampled visible nodes define a scene for the context (Task 4). \label{fig:randomscenegeneration}}
\end{figure}

In this task, we demonstrate how we can use another generative ability of our Triway BM: we can select a hidden node (or more of them, leaving the other hidden neurons randomly initialized or set to zero), and sample visible nodes (including relations) that describes a scene. Figure \ref{fig:randomscenegeneration} shows some visual examples.  

\section{Conclusion}
We have proposed a novel method based on Boltzmann Machines for contextualized scene modeling. For this end, we extended BM by adding spatial relations between objects that are shared across different objects in the scene. Compared to RBM and DBM, we show that our hybrid model performs better in several scene analysis and reasoning tasks, such as finding relations, missing objects and out-of-context objects. Moreover, being generative, our model allows generating new scenes given a context or a part of the scene (as a set of objects).

\section*{Acknowledgment}

This work was supported by the Scientific and Technological Research Council of Turkey (T\"UB\.{I}TAK) through project called ``Context in Robots'' (project no 215E133). We gratefully acknowledge the support of NVIDIA Corporation with the donation of the Tesla K40 GPU used for this research.

\bibliographystyle{IEEEtran}
\bibliography{references}

\addtolength{\textheight}{-12cm}

\end{document}